# SUPPORT VECTOR MACHINE FOR HANDWRITTEN CHARACTER RECOGNITION

**Jomy John**[1],

[1]Department of Computer Science, K. K. T. M. Government College, Pullut.

**Abstract:**

Handwriting recognition has been one of the most fascinating and challenging research areas in field of image processing and pattern recognition. It contributes enormously to the improvement of automation process. In this paper, a system for recognition of unconstrained handwritten Malayalam characters is proposed. A database of 10,000 character samples of 44 basic Malayalam characters is used in this work. A discriminate feature set of 64 local and 4 global features are used to train and test SVM classifier and achieved 92.24% accuracy.

## INTRODUCTION

Handwriting recognition has been one of the most fascinating and challenging research areas of image processing and pattern recognition in the recent years. It contributes enormously to the improvement of automation process and upgrades the interface between man and machine in numerous applications which include, reading aid for blind, library cataloguing, ledgering, processing of forms, cheques and faxes and conversion of any handwritten document into editable text etc. As a result, the off-line handwriting recognition continues to be an active area of research towards exploring the innovative techniques to produce adequate accuracy. Even though, sufficient studies have performed in foreign scripts like Chinese, Japanese and Arabic characters, only few works can be traced for handwritten character recognition of Indian scripts. Majority of them were based on Bangla and Devnagiri script. India is a multi lingual multi script country with twenty two scheduled languages and Malayalam in one among them. In Indian language scripts, the concept of upper case and lower-case characters is not present. Most of the Indian languages are derived from Ancient Brahmi and are phonetic in nature and hence writing maps sounds of alphabets to specific shapes. A detailed description of the works on South Indian scripts can be read in [1]. John et al. [2] have proposed a method that uses chain code histogram and image centroid for the purpose of extracting features and a two layer feed forward network with scaled conjugate gradient for classification. In another work, CBIR based retrieval of similar Malayalam characters are described [3]. This paper deals with recognition of handwritten basic Malayalam characters using support vector machine classifier.

The paper is organized as follows: Materials and Methods section is covered in Section II; Experimental setup and classification results are discussed in Section III; Section IV concludes the paper.

1. **MATERIALS AND METHODS**

Handwriting recognition in general, undergoes through pre-processing phase, feature extraction phase and classification phase.

A. *The pre-processing phase*

Data was collected from different persons of the population in Kerala, including different age groups and different educational levels without imposing any constraints. It represents wide variety of writing styles. Digitization of collected samples are done by a Flat-bed scanner (manufactured by HP, Model Name: Scanjet 2400), by setting dpi to 300. Neatly written character samples from each class is displayed in Fig. 1

1) *Noise Removal and Binarization*

Noise is defined as any degradation

*Fig. 1 Samples from each basic character of Malayalam Language*

SCIENCE

in the image due to external disturbance. Quality of handwritten documents depends on various factors including quality of paper, aging of documents, quality of pen, color of ink etc. A median filter is used to remove unwanted noise. In this paper a median filter with a 3x3 mask is applied. Binarization is required to concentrate more on the shape of the characters and remove background details from the objects. Thresholding is the simplest way of binarization. Otsu's method [4] uses global thresholding method which has proved to perform best on average. In this paper classic Otsu method is used for binarization.

*2) Skew detection and correction*

Digitization using scanner may lead to skew in the image. During preprocessing stage the image file is checked for skewing. Skew with respect to page border is detected and corrected by rotating the image by the skew angle in the opposite direction. The function for skew detection checks for an angle of orientation between ± 15 degrees and if detected then a simple image rotation is carried out till the lines match with horizontal border.

*3) Segmentation*

Horizontal histogram profile, which is a running count of the black pixels in each row, is calculated and the minimum value between each peak values are used for line segmentation. From each line containing characters, character segmentation starts with connected component labeling. The minimum bounded rectangle containing the component are extracted and stored in the database. It contains broken and distorted characters also.

*4) Normalization and Thinning*

It is the process of converting the random sized image into standard sized image. This, size normalization avoids inter class variation among characters. A bicubic interpolation technique is used here to convert each image into 32x32, where the output pixel value is a weighted average of pixels in the nearest 4x4 neighborhood. Thinning is an image preprocessing operation performed to make the image crisper by reducing the binary-valued image regions to lines that approximate the skeletons of the region. In this paper, thinning is done by morphological thinning algorithm. Samples of normalized, thinned, binary character images are displayed in Fig. 2

*B. Feature Extraction*

Features represent salient information of the input image. There are different feature extraction methods proposed for character images. Since Malayalam characters are rich in shapes, each character is represented by means of a vector of features. The feature extractor, fed with the pre-processed character image generates local and global features. In order to extract local features, the character image is divided into cells of equal size by using an arbitrary grid, as shown in Fig. 3. If the preprocessed character image is A, where $A = (a_{ij}) \in$, $Z = \{0,$ then, the local feature vector V∈RM x N of A is given by

$$V = [v_1 \ v_2 \ v_3 \ldots v_n]$$

$$V_k = \sum_{i=l_k}^{u_k} \sum_{j=l_k}^{u_k} (a_{ij})$$

where $l_k$ and $u_k$ are the lower bound and upper bound of cell k, 1≤k≤n.

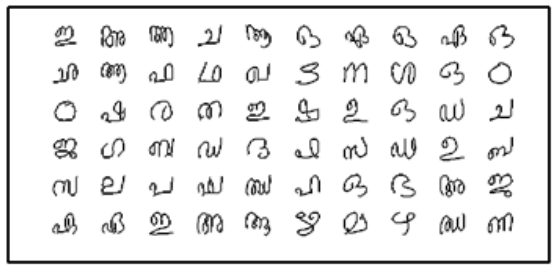

*Fig.2 Samples of normalized thinned binary image*

Then the character image അ is used to extract four global features. The first global feature is the ratio of the width and the height of the image (w/h,) with respect to the original segmented image. The remaining global features are being the number of endpoints, number of cross points and the number of branch points of thinned image. These four global features are added with the local feature set to give an idea about the entire shape of the handwritten character. The number of local features can be arbitrarily determined by changing the number of cells by varying the grid size. The feature selection technique is chosen in order to keep the feature number as low as possible. Therefore the feature set was tested by setting the size of grid as 16 × 16, 8 × 8, 4 × 4 and 2 × 2, giving 4, 16, 64 and 256 features. The one giving the best results (4 × 4) was selected. In the reported experiments a feature vector having 68 elements are used. Four features are global (width/height ratio, number of endpoints, number of cross points and number of branch points) while the remaining 64 are generated from 64 cells, placed on a regular 4×4 grid.

*C. Classification using SVM*

SVM is one of the popular techniques for pattern recognition and is considered to be the state-of-the-art tool for linear and non-linear classification [5]. It has been found to have empirically good performance in solving many application problems. It belongs to the class of supervised learning algorithms, based on statistical learning theory.

The SVM classifier has been proposed for binary classification in literature and learning algorithm comes from an optimal separating hyper-plane, developed by Vapnik [5]. For binary classification, a linear decision function f(x) is used: f(x)= $w^T x+b$ where w is the weight vector, b is a bias. Classification is given by sign of f(x). It can be -1 or +1. The optimal solution is obtained when this hyper plane is located in the middle of two classes (Fig. 4) and the points that constrain the width of the margin are called support vectors. The distance between two classes can be expressed by $2/\|w\|$ In the case of linearly inseparable data; it is difficult to determine a linear decision function. So, the original pattern space is mapped into a high dimensional feature space through some non-linear mapping functions. This option maps non-linear problem in low dimensional space into a linear problem in the high dimensional space, so that the optimal separating hyper plane can be constructed. Given a training set of instance – label pair $(x_i, y_i)$, i=1,…,l, where $x_i \in R_n$ and y∈{1, -1}, the support vector require the solution of the following optimization problem [5].

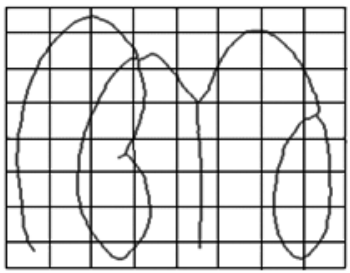

**Fig. 3** Grid covering the thinned image of character (അ)

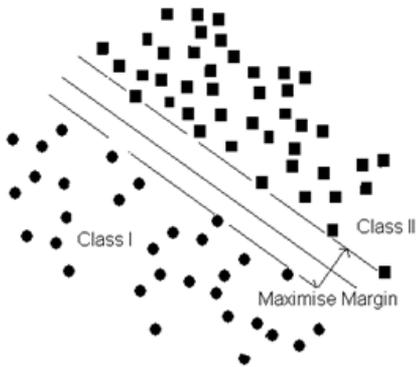

Fig. 4 *Maximum margin hyper plane separating 2 classes*

$$\min_{w,b,\xi} \frac{1}{2} w^T w + C \sum_{i=1}^{l} \xi_i$$

subject to
$$(y_i\, w^T \phi(X_i) + b) \geq 1 - \xi_i$$
$$\xi_i \geq 0, i = 1, …, l$$

Here training vectors $X_i$ are mapped into a higher dimensional space by the function $\phi$. Support Vector Machine finds a linear separating hyper plane with maximal margin in the respective mapped space. In practical problems, it is not possible to find out a hard margin hyper plane as classes may overlap due to noise. Therefore $\xi_i$ is introduced as a slack variable to relax the constraints and b is the bias term. The Kernel function is termed as:
$$K(X_i, X_j) = \phi(X_i)^T \cdot \phi(X_j)$$

Basic kernel functions include:

Linear: $K(X_i, X_j) = x_i^T \cdot x_j$

Polynomial:
$$K(X_i, X_j) = (x_i^T \cdot x_j + 1)^d$$

Radial basis function:
$$K(X_i, X_j) = \exp(-ã\|x_i - x_j\|^2),\ \gamma > 0$$

Sigmoid:
$$K(X_i, X_j) = \tanh(ăx_i^T \cdot x_j + r)$$

where γ, r and d are kernel parameters. The effectiveness of SVM depends on the selection of kernel, the kernel's parameters,

SCIENCE

and soft margin parameter C.

*1. Multi class SVM*

The binary SVM can be extended to multiclass [6]. Multiclass SVMs are usually implemented by combining several two-class SVMs. Two approaches common in practice are one-versus-all method and one-versus-one method. The one-versus-all method represents the earliest and most common multiclass approach used for SVMs. Each class is trained against the remaining N-1 classes that have been collected together. The winner-takes-all strategy is used for final decision. The winning class corresponds to the class with highest output function. For one classification N binary classifiers are needed. In one-versus-one method, each one is trained on data from two classes. While testing for each class, the max-wins voting strategy is used to classify an unknown sample. For one classification, N (N-1)/2 binary classifiers are needed

III. **RESULT AND DISCUSSIONS**

The experiment had been carried out on a database of 10,000 handwritten isolated Malayalam characters written by 228 different writers. It contains all the 44 basic characters. In order to represent each character as a vector of features, different sized grids namely, $16 \times 16$, $8 \times 8$, $4 \times 4$ and $2 \times 2$, are placed on pre processed image, giving 4, 16, 64 and 256 cells. This yields 4, 16, 64 and 256 local features based on the occurrence of pixels in each cell. Along with these local features, four global features, namely, width/height ratio, number of endpoints, number of cross points and number of branch points are added. The width/height ratio is calculated from initial segmented image, while the remaining three are calculated from thinned image. This represents the feature sets having 8, 20, 68 and 260 features.

Support vector machine with polynomial and radius basis function

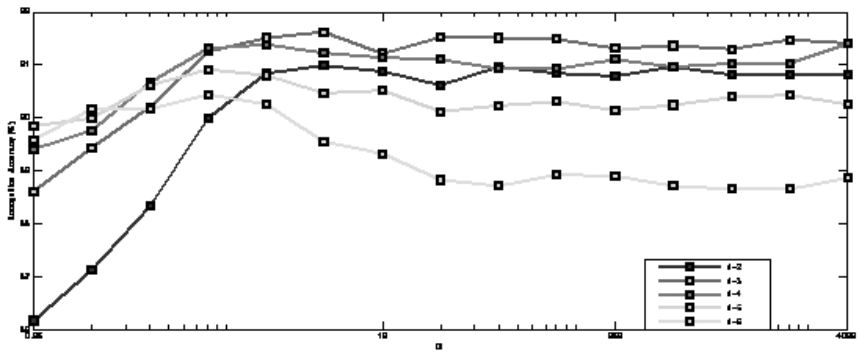

**Fig. 5** Performance plot of polynomial kernel for d=[2, 3, 4, 5, 8]

SCIENCE

(RBF) kernels have reported superior performance in pattern recognition applications. Accordingly, both have been applied in this study. For our support vector machine, one-versus-all method is used to construct 44 binary classifiers. That is, each classifier was constructed by separating one class from the rest. Classification result was made by choosing the class with maximum output value. SVMlight, an implementation of Vapnik's Support Vector Machine [7] is used in this work. The optimization algorithms used in SVMlight are described in Joachims [7]. For RBF kernel, the accuracy is estimated using various parameters for $\gamma=[2^4, 2^3, 2^2, \ldots, 2^{-10}]$ and $C=[2^{12}, 2^{11}, 2^{10}, \ldots, 2^{-2}]$. For polynomial kernel, the parameter d has only few choices, $d \in \{2\ 3\ 4\ 5\ 6\}$. Cross validation strategy is used here to find out optimal hyper parameters. 10-fold cross validation technique is used, in which training data are randomly divided into 10 equal sized subsets and SVM classifier is trained using 9 subsets and tested using the remaining subset. Training is repeated 10 times and the total recognition rate for all the 10 subsets that are not included in the training data is calculated. Experiments have carried out with feature sets of 8, 20, 68 and 260 features. For all experiments, the database is split with a random process into training (80%) and testing (20%). From the preliminary stage itself it was trivial that the one giving the best performance was based on 4 × 4 grid (68 features). So, further experiments are carried out only with respect to this feature set. In order to avoid bias towards a particular choice of training and testing data, the experiment is repeated five times and the average test accuracy is reported. Performance plot of polynomial kernel for various parameters of C and d are shown in Fig. 5. Polynomial kernel with parameter 3 outperformed the others for C≥2. The optimal accuracy is obtained for C=8. The highest recognition accuracy with RBF kernel is obtained for C=64, γ =0.02. The plot by fixing C as 64 and by varying γ is displayed in Fig. 6. The classification result is tabulated in Table 1. Error rate of each individual character is depicted in Table 2. The lowest error rate was obtained for character ഠ and the error rate was comparatively high for (ഉ, ഇ, ള), ( എ,ഐ), (ഡ,ഢ), (ഞ,ണ) and (ഭ, ഋ) as they have cross resemblance with each other and this affected the overall accuracy.

## IV. CONCLUSION

Malayalam handwritten character recognition system using SVM have been proposed in this paper. Discriminative information from the characters is extracted by using a grid of size 4 × 4. Each character is represented by means of a vector of 68 features. SVM with RBF and polynomial kernel have used for classification. The highest classification accuracy of 92.24% is obtained for RBF kernel with parameter 0.02. As our database contains samples of characters that are even undistinguishable for human, the accuracy reported is acceptable. Enhanced classification accuracy can be achieved if more robust features are included to reduce misclassified characters or by using a multi stage classifier. Future work can incorporate these techniques to attain improved result.

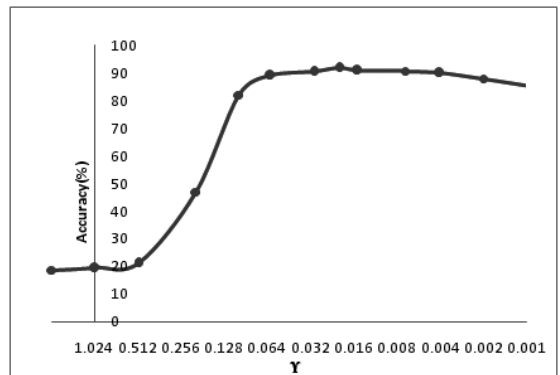

**Fig. 6** Performance plot of RBF kernel for various

**Table 1:** Classification result for RBF and Polynomial kernels

| Data Number | Kernel | Classification Accuracy (%) | | | | | Average Classification Accuracy (%) |
|---|---|---|---|---|---|---|---|
| | | Iteration 1 | Iteration2 | Iteration 3 | Iteration 4 | Iteration 5 | |
| 10,000 | RBF C=64, γ=.02 | 92.0500 | 92.4500 | 92.3500 | 92.2000 | 92.1500 | 92.24 |
| | Poly C=8, d=3 | 90.4000 | 91.4000 | 90.0500 | 90.7000 | 91.6000 | 90.83 |

**Table 2:** Error rate of each individual character using RBF kernel

| Class | Char | Error rate | Class | Char | Error rate | Class | Char | Error rate | Class | Char | Error rate |
|---|---|---|---|---|---|---|---|---|---|---|---|
| 1 | അ | .0045 | 12 | ഏ | .0055 | 23 | ണ | .008 | 34 | യ | .0035 |
| 2 | ആ | .0025 | 13 | ഒ | .0035 | 24 | ത | .0035 | 35 | ര | .0015 |
| 3 | ഇ | .004 | 14 | ച | .003 | 25 | ഥ | .0035 | 36 | ല | .004 |
| 4 | ഉ | .009 | 15 | ഛ | .001 | 26 | ദ | .0065 | 37 | വ | .0045 |
| 5 | ഋ | .0015 | 16 | ജ | .002 | 27 | ധ | .004 | 38 | ശ | .0025 |
| 6 | എ | .0065 | 17 | ഝ | .005 | 28 | ന | .0035 | 39 | ഷ | .0055 |
| 7 | ഏ | .0055 | 18 | ഞ | .0065 | 29 | പ | .002 | 40 | സ | .0055 |
| 8 | ഒ | .0055 | 19 | ട | .0025 | 30 | ഫ | .005 | 41 | ഹ | .002 |
| 9 | ക | .003 | 20 | ഠ | .003 | 31 | ബ | .006 | 42 | ള | .0075 |
| 10 | ഖ | .0045 | 21 | ഡ | .008 | 32 | ഭ | .007 | 43 | ഴ | .0045 |
| 11 | ഗ | .0015 | 22 | ഢ | .007 | 33 | മ | .0035 | 44 | റ | .00 |